\documentclass[letterpaper]{article}
\usepackage{aaai20}
\usepackage{times}
\usepackage{helvet}
\usepackage{courier}
\usepackage{amsmath}
\usepackage{xspace}
\usepackage{xcolor}
\usepackage{graphicx}
\usepackage{amssymb}
\usepackage{subcaption}
\nocopyright 

\newcommand{\riker}{\textsc{Riker}\xspace}
\newcommand{\pretcil}{\textsc{PReTCIL}\xspace}
\newcommand{\doctype}{paper\xspace}

\newcommand{\citet}[1]{\citeauthor{#1} \shortcite{#1}} \newcommand{\citep}{\cite} 

\newtheorem{definition}{Definition}

\begin{document}
%
\title{Helpfulness as a Key Metric of Human-Robot Collaboration}
\author{Richard G. Freedman\\
SIFT, LLC\\
319 N. 1st Ave., Suite 400, Minneapolis, MN 55401\\
rfreedman@sift.net
\And
Steven J. Levine\\
MIT CSAIL\\
32 Vassar St., Cambridge, MA 02139\\
sjlevine@alum.mit.edu
\AND
Brian C. Williams\\
MIT CSAIL\\
32 Vassar St., Cambridge, MA 02139\\
williams@csail.mit.edu
\And
Shlomo Zilberstein\\
University of Massachusetts Amherst\\
140 Governors Dr., Amherst, MA 01003\\
shlomo@cs.umass.edu
}
\maketitle
\begin{abstract}
\begin{quote}

As robotic teammates become more common in society,
people will assess the robots' roles in their interactions along many dimensions. 
One such dimension is effectiveness: people will ask whether their robotic partners are trustworthy and effective collaborators.
This begs a crucial question: how can we quantitatively measure the helpfulness of a robotic partner for a given task at hand? 
This \doctype seeks to answer this question with regards to the interactive robot's decision making.  We describe a clear, concise, and task-oriented metric applicable to many different planning and execution paradigms. The proposed helpfulness metric is fundamental to assessing the benefit that a partner has on a team for a given task. In this \doctype, we define helpfulness, illustrate it on concrete examples from a variety of domains, discuss its properties and ramifications for planning interactions with humans, and present preliminary results. 
\end{quote}
\end{abstract}

\section{Introduction\label{intro}}

Interactions between decentralized, autonomous agents can yield a range of positive to negative experiences.  Positive interactive experiences can establish or reinforce relationships of trust, but negative ones may require explanations to justify why trust should be maintained.  A common explanation 
is, ``I was trying to \emph{help}.''  The notion of \emph{helpfulness} is that an agent is, with honest intentions, trying to play a positive role with the task at hand. However, this term is vague when considered as a justification of behavior---what motivated the choice as a helpful one, and did it actually help?

Furthermore, as robots become more tightly involved in interactions with humans, people will prefer robots that are good interaction partners. Trust and helpfulness are two criteria that contribute to evaluating the quality of interaction partners, and they are coupled.  A robot that successfully helps someone can increase their trust in the robot.  Likewise, a person who trusts a robot is more likely to accept that robot's attempts to help rather than avoid it.

Helping does not typically have an explicit definition like most artificial intelligence planning problems.  In particular, we cannot formulate goal criteria directly unless other agents' goals are known.
It is thus \emph{dynamic as other agents' goals change}.  Without a clear definition for how intelligent interactive robots can help, it is important to identify important factors that contribute to helpfulness and use those factors to create a general-purpose metric.

This \doctype provides an initial proposal towards such a metric for quantitatively measuring a robot's helpfulness with respect to artificial intelligence decision making.  Section~\ref{preview} introduces the intuition behind helpfulness, followed by related work in Section~\ref{related}.  We then present more formal definitions of helpfulness in Section~\ref{defn} and their use within planning and execution in Sections~\ref{Sec:Cost},~\ref{Sec:ExecutionRisk}, and~\ref{planningEx}.  Preliminary experimental results are described in Section~\ref{results.exp}. We conclude with a discussion of open questions and challenges that our proposed definition introduces in Section~\ref{conclusions}.

\section{Helpfulness in a Nutshell\label{preview}}

\begin{figure}
    \centering
    \includegraphics[width=\linewidth]{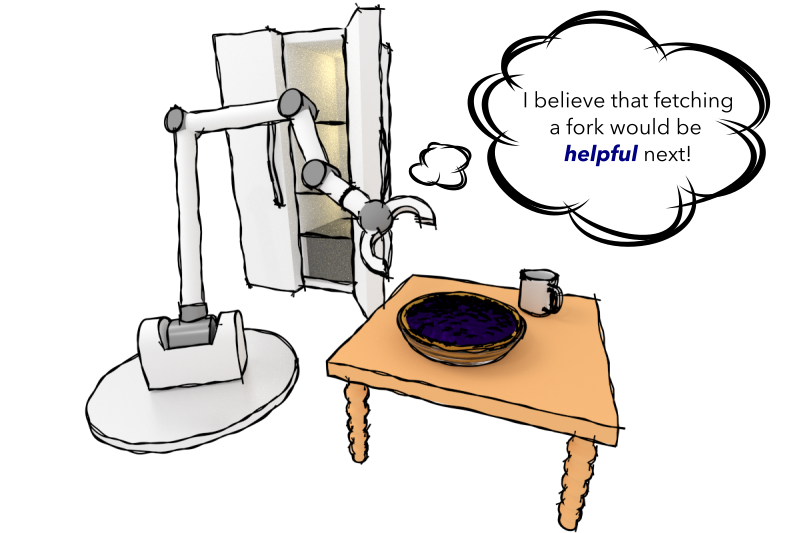}
    \caption{A robot's plan is \emph{helpful} if it decreases the effort for the team to complete a task, compared to the human performing the task alone. If the task is to prepare a blueberry pie, an action in a helpful plan may be to fetch the fork.}
\label{fig:HelpfulCookingRobot1}
\end{figure}

This section serves to illustrate the key intuitions of helpfulness via a grounded example. Consider a household robot assisting a human with a cooking task in the kitchen as shown in Figure~\ref{fig:HelpfulCookingRobot1}. Suppose the human takes some blueberries out of the refrigerator. Based on previous mornings, the robot responsively infers that their goal is to make a blueberry pie. Given that goal, what actions should the robot take to assist with it?
Preheating the oven or fetching other supplies (such as the pie plate) intuitively seem helpful because they bring the team closer to completing the goal.

\begin{figure}
    \centering
    \includegraphics[width=\linewidth]{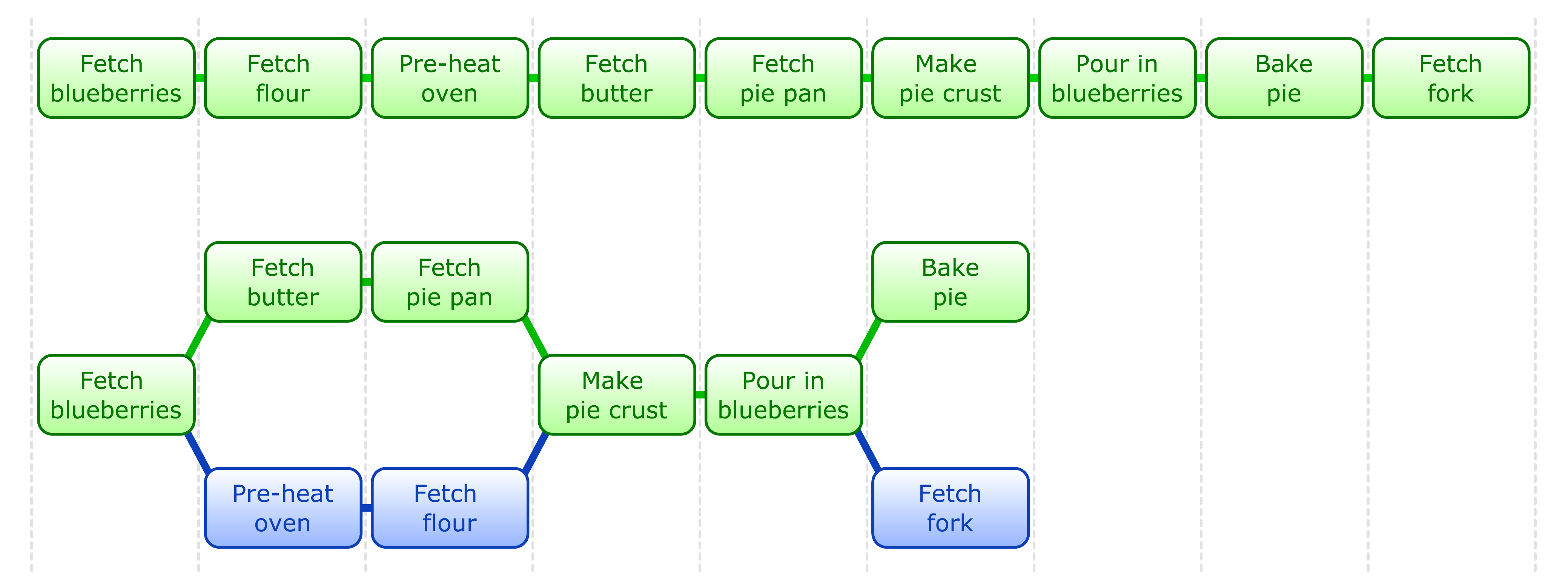}
    \caption{Top: plan for a human to make a blueberry pie alone, which takes 9 steps. Bottom: plan for a human-robot team to make a blueberry pie, taking 6 steps. Human actions are shown in green, and robot actions are shown in blue.}
\label{fig:CookingPlans}
\end{figure}

Figure~\ref{fig:CookingPlans} illustrates two example plans that accomplish the goal. On top, we see a plan for a single agent (the human) to prepare the blueberry pie. This plan has nine steps, and only the human performs each action. On bottom, we see a plan for a human-robot team preparing the same blueberry pie. During certain steps, the robot can act in parallel with the human, thus saving time. For instance, the robot can preheat the oven at the same time that the human fetches the butter. As a result, the human-robot team takes only six steps to make the blueberry pie.

Intuitively, the robot is very helpful in this example.  It selects intelligent actions that align well with the task at hand. Quantitatively, this is reflected by the comparison of the overall \emph{cost} of executing the plan for the human-robot team versus the agent alone:
\begin{align*}
    cost(\text{human alone}) &= 9 \\
    cost(\text{human-robot team}) &= 6
\end{align*}

In this example, the cost is the total number of steps until achieving the goal of being ready to eat the blueberry pie. \citet{DBLP:conf/aaai/FreedmanZ17} referred to the difference between these two costs---human-alone versus human-robot team---as \emph{helpfulness} $H$, which 
quantifies the benefit gained from adding this particular robot to the team.

We further introduce \emph{relative helpfulness}, 
which intuitively moves from absolute costs to a ratio of costs:

\begin{align*}
    H_R = \frac{cost(\text{human alone}) - cost(\text{human-robot team})}{cost(\text{human alone})}
\end{align*}

By dividing by the cost of the human alone, we can measure the benefit of adding the robot as a percentage. Taking the blueberry pie example, the relative helpfulness is $3/9$, indicating that adding this robot to the team decreased the cost by 33.3\%. In this sense, relative helpfulness can be seen as quantifying the improvement in team efficiency.

Thus far, we have only discussed the case where cost is the number of steps to reach the goal. However, it is important to note that we can use any definition of cost, depending on the domain. Section~\ref{Sec:Cost} describes other possible $cost(\cdot)$ functions.

Regardless of cost metric, there is a wide range of possible helpfulness values. In the limit, imagine we have an extremely helpful robot that is capable of achieving the entire task with no effort. For example, suppose the robot already had a blueberry pie prepared from earlier.  Hence adding the robot to the team means that no actions are needed to achieve the goal. In this extreme case, $cost(\text{human-robot team}) = 0$ while $cost(\text{human alone}) = k > 0$. The relative helpfulness is therefore $1$, indicating a perfectly helpful robot. In contrast, suppose we have a robot that is incapable of assisting or takes ``no-operator'' (no-op) actions (choosing to do nothing) at every step, or selects actions that are irrelevant to the task at hand (such as fetching spaghetti and a pot to boil water). In this case, the cost of the human performing the task alone will be equal to the cost of the team, resulting in a helpfulness value of $0$. Thus, for typical collaborative interactions between the human and the robot, relative helpfulness values lie in the range of $[0, 1].$

Finally, we wish to highlight one key property of helpfulness: it is a task-oriented metric. Depending on the team's goals, actions may or may not contribute to a helpful execution. In the case of making a blueberry pie, a robot that begins boiling water would not be very helpful. However, if the goal is to make spaghetti, it would be a very helpful action. Thus, it is important to keep in mind that helpfulness is a function of the team's goals.

\subsection{Special Cases of Helpfulness\label{preview.specialcases}}
It is possible for helpfulness and each of its variations to take on atypical values.  First, helpfulness can be negative.  This usually occurs when a robot has an adversarial role in the interaction, or it can occur when a well-meaning robot happens to be accident-prone. In the case of the kitchen scenario, a clumsy robot that consistently drops items---flour, glass, pie plates, etc.---may actually cause the team to incur a higher cost compared to the human performing the task alone. Then $cost(\text{human-alone}) < cost(\text{human-robot team})$, which evaluates to $H < 0$.

\begin{figure}
    \centering
    \includegraphics[scale=0.32]{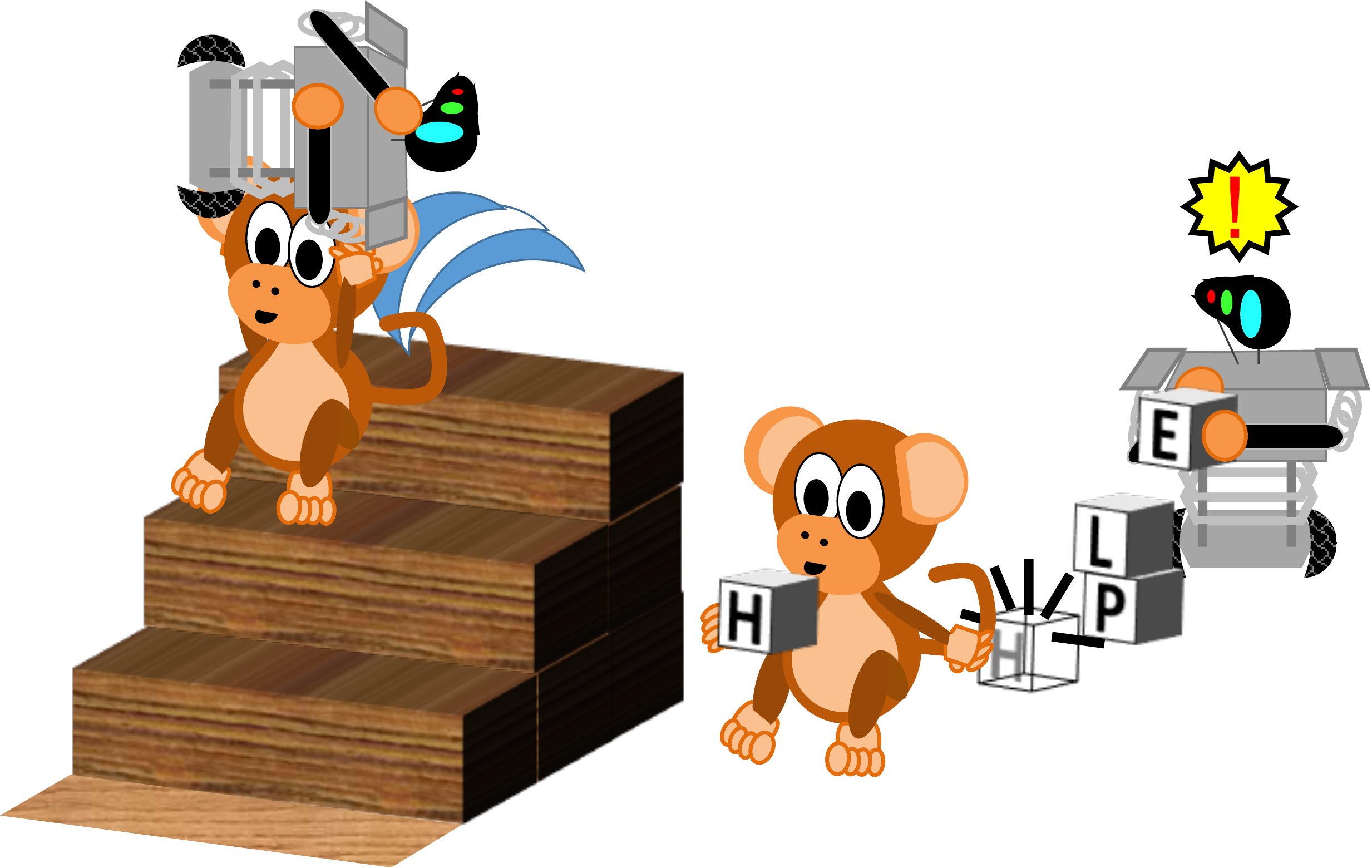}
    \caption{(Left) The monkey's helpfulness evaluates to $\infty$ because the robot cannot go down the stairs by itself.  (Right) The monkey's helpfulness evaluates to $-\infty$ because the block they took is required for the robot to achieve its goal of stacking blocks to spell ``HELP''.}
    \label{fig:infiniteHelpfulnessExamples}
\end{figure}

Second, helpfulness can reach the extremes of infinity.  In automated planning, a cost of $\infty$ implies that there is no solution---the effort to solve the problem is never-ending.  When $cost(\text{human-alone})$ is a finite number and $cost(\text{human-robot team}) = \infty$, then the interacting robot makes reaching the goal impossible (e.g., an adversarial kitchen robot that cuts the power cord to the oven) and the helpfulness is $H = -\infty$.  When  $cost(\text{human-alone}) = \infty$ and $cost(\text{human-robot team})$ is a finite number, then the interacting robot is the only reason that the human can achieve the goal (e.g., a human who is less adept at baking might always burn the blueberry pie while the assistive kitchen robot never burns it) and the helpfulness is $H = \infty$.  Figure~\ref{fig:infiniteHelpfulnessExamples} illustrates cases for each extreme.

\section{Related Work\label{related}}

The general notion of evaluating interactions between humans and robots has been an ongoing research topic.  The majority of the metrics for task-oriented interactions focus on various performance aspects with respect to the human, robot, or entire team \cite{common-hri-metrics}.  For measuring robot performance within an environment where they may interact with humans, there is typically a greater focus on autonomy.  A robot that properly assesses when and how to interact with people will enable them to do their job with fewer interruptions and waste less time overseeing the robots.  Six such metrics include task effectiveness, neglect tolerance (the robot's performance over time without human involvement), robot attention demand, free time, fan-out (how well a human can work with multiple robots at once), and interaction effort (amount of human involvement during interaction)~\cite{six-metrics-hri}. Cognitive metrics for team collaboration, such as the similarity of team members' mental models, have also been considered~\cite{pina2008identifying}.

In addition to evaluating performance as a post-interaction assessment, \citeauthor{hrt-metrics-survey} \shortcite{hrt-metrics-survey} consider a variety of performance metrics for constructing effective human-robot teams.  Their survey of existing metrics spans beyond the robot's autonomy to further include system-wide evaluations for the human-robot team as a whole.

One metric often used to measure the performance of a team is the overall makespan, or the \emph{time the team takes to complete the task}~\cite{CrandallEfficiencyInInteraction,ChaskiThesis}. With increasing helpful involvement from the robot, this time should decrease since the robot will do more work in parallel with the human. 
Another related metric is that of \emph{human idle time}~\cite{ChaskiThesis}. Intuitively, teams are more efficient when members are less idle; the team members make progress towards their goal by taking action. Sometimes, one teammate is ``blocked'' by another teammate --- i.e., one teammate's action has preconditions that another teammate's action achieves. For example, the human may not be able to make the pie crust until the robot fetches a required ingredient such as flour. If one agent is forced to wait for another teammate, the teamwork is less effective. Hence, a helpful robot may choose to unblock the actions earlier (e.g., the robot may fetch the flour earlier in the plan if possible) so as to improve the amount of parallelism of the team and their overall efficiency at completing the task.

Focusing on the robot within a team, it is sometimes useful to measure the \emph{execution latency} of a robot as it makes decisions. Intuitively, this is the time the robot requires to perform its computations and act after observing the human's actions. Lower execution latency is preferred because the interaction feels more fluid when the robot acts quickly instead of with pauses. Execution latency was previously evaluated in the contexts of temporal plan executives~\cite{DrakeJAIR} as well as human-robot interaction~\cite{ChaskiICAPS,LevineJAIR,RikerLevinePhdThesis}.

Another metric identified to correlate with team performance is the \emph{amount and type of communication} between team members. A robot that requires constant, explicit verbal communication (i.e., being commanded what to do) is likely to be less helpful than a robot that infers intent through implicit communication (i.e., observing non-verbal human cues). This may be related to the \emph{switching cost}, where the human must pause the current task in order to assist the robot with explicitly-communicated guidance~\cite{ChaskiThesis}.

Finally, qualitative, holistic metrics can be very important metrics for assessing the quality of human-robot interaction. Assessing the degree to which humans agree or disagree with statements such as ``I trust the robot'' on a Likert scale have successfully been used in prior work~\cite{ChaskiThesis}.  With respect to humans' perspective on the potential role of robots on cooperative teams, \citet{Gombolay-RSS-14} observed through both human-robot assembly experiments as well as surveys that people are fine giving up their autonomy on scheduling and role allocation for tasks if it leads to greater task efficiency.  In a similar study, \citet{efficient-hri-humanlikehelp} found similar preferences when a robot decided whether to help people with their tasks mid-execution.  Furthermore, they found that people preferred to have some control over telling the robot when to help as well as having the robot be more proactive to offer assistance.

The relationship between being helpful and improving efficiency has also been identified within psychology.  An act as simple as holding the door for others not only reduces the effort they need to enter/exit a building, but those individuals also show signs of hurrying their pace to reach the door \cite{doorholding_PsychSci}.  This responsive behavior implies intent to reduce the time that the person needs to hold the door open, reciprocating the improvement of efficiency.  Independently of this research, the notion of helpfulness was quantified as the reduction in an agent's effort to perform a task when another agent participates \cite{DBLP:conf/aaai/FreedmanZ17,freedmanthesis}.  The work in this \doctype builds upon this quantification to consider variations for additional informative metrics and specific applications.

Although the former work evaluated the helpfulness of an agent that integrated automated planning with plan and intent recognition, the scope of this \doctype will not specifically cover the impact of recognition algorithms on helpfulness in human-robot interaction.  There is certainly a close relationship however, as understanding what others are doing and why are essential to knowing who needs help and in what ways.  Previous approaches to recognizing people's intents in interactions include the theory of mind \cite{alami-tom-hri}, parsing grammars that capture task hierarchies \cite{gwmkcp_aiide16}, and epistemic reasoning \cite{shvo_aamas20}.

\section{Helpfulness Defined\label{defn}}
A common output of decision making algorithms is a \emph{plan} $\pi$, which is a sequence of actions.  Ideally, $\pi$ is a solution to a problem such that an agent can, from a specified initial state, perform the actions in order and reach a state that satisfies the problem's goal conditions $G$.  If the solution is optimal, then no other plan can solve the problem with lower cost; we call the optimal plan $\pi^{*}$.

We will refer to the human and robot agents as 
$A_{H}$ and $A_{R}$ respectively.  We assume that $A_{H}$ has some internal goal towards which they plan and act, and $A_{R}$ is the interactive partner who intends to help $A_{H}$ (regardless of whether $A_{R}$ knows $A_{H}$'s goals ahead of time or infers them through observation). 
Collaboration prior to the interaction is ideal to facilitate explaining intentions and establishing trust between the agents, but these preparations may not always be possible.  We consider the impacts of their (de)centralized plans when acting together.

In this case, $\pi_{A_{H}}$ and $\pi_{A_{R}}$ are the agents' independent plans where they act on their own---if $A_{H}$'s plan is unknown, then we assume the agent is optimal and use $\pi_{A_{H}}^{*}$ for a lower bound helpfulness estimate.  As the agents act in the world and observe each other's actions, then they can adjust their plan and predict how other agents will interact. 
These responsive plans \cite{DBLP:conf/aaai/FreedmanZ17} have the form $\pi_{A_{R} \rightarrow A_{H}}$ such that $A_{R}$ is performing the plan with respect to $A_{H}$'s interaction.  As the agents continue to interact, they may recursively shape their plans to have forms such as $\pi_{A_{i \in \left\lbrace R,H \right\rbrace} \rightarrow \cdots \rightarrow A_{R} \rightarrow A_{H}}$.  Alternatively, agents can have a centralized joint plan $\pi_{A_R + A_H}$ that dictates how they will work together, which \emph{assumes a shared and known goal}.  We assume $cost\left(\pi\right) = \infty$ if $\pi$ does not exist.

\begin{definition} The \emph{helpfulness} of agent $A_{R}$ in a joint plan $\pi_{A_{R} + A_{H}}$ is the change in cost from agent $A_{H}$ acting on their own to both agents working simultaneously 
\begin{align*}
H\left(A_{R}, \pi_{A_{R} + A_{H}}\right) = cost\left(\pi_{A_{H}}\right) - cost\left(\pi_{A_{R} + A_{H}}\right) . 
\end{align*}
\end{definition}

\begin{definition} The \emph{helpfulness} of agent $A_{R}$ in a responsive plan $\pi_{A_{R} \rightarrow A_{H}}$ is the change in cost from agent $A_{H}$ acting on their own to both agents working simultaneously 
\begin{align*}
H\left(A_{R}, \pi_{A_{R} \rightarrow A_{H}}\right) = & \\ cost\left(\pi_{A_{H}}\right) & - cost\left(\pi_{A_{H} \rightarrow A_{R} \rightarrow A_{H}}\right) .
\end{align*}
\end{definition}

Although these helpfulness metrics evaluate an agent on a plan-by-plan basis, it is difficult to use this for comparison purposes.  Assessing an agent's overall helpfulness between multiple plans and problems, and furthermore comparing agents who have different planning paradigms, poses a risk of unfair bias when the measured plans all have different typical costs.  Specifically, there are far fewer opportunities to be helpful when responding to a plan of cost $c$ compared to responding to a plan of cost $k \cdot c$ for some $k >> 1$.

The difference in cost does not reflect \emph{the potential to be helpful}, which matters in situations where $A_{H}$ can decide whether to perform the task and they are aware of $A_{R}$'s intentions to interact---in this case, helpfulness can serve like value of information.  As a grounded example, consider the two tasks where
\begin{itemize}
\item $A_{H}$ must move a large piece of furniture across a room \cite{Mortl:2012:RRP:2421541.2421549} and
\item $A_{H}$ must move all furniture in a room onto a truck.
\end{itemize}
The first task is inherently lower-cost so that $A_{H}$ can consider moving the piece of furniture alone, and any help $A_{R}$ provides is a nice bonus.  The second task is much more costly to perform and $A_{H}$ will likely want to ensure that $A_{R}$ will provide sufficient help \emph{relative to the task's effort} before starting to execute a plan.  This can extend to waiting for $A_{R}$ to be ready to interact in a more helpful manner \cite{Zaaai15_semiautonomouSystBluesky} or finding another interactive partner who is available to provide sufficient help \cite{Rosenthal_CobotAskHelp}.

We thus propose normalizing helpfulness to account for how much help $A_{R}$ could provide in the first place:
\begin{definition}\label{def:respPlan.helpfulness.challenges.normhelpfulness}
The \emph{normalized helpfulness} of agent $A_{R}$ in a plan $\pi$ is the ratio comparing the helpfulness $H\left(A_{R}, \pi\right)$ to the best-case helpfulness where $A_{H}$ and $A_{R}$ worked together simultaneously from the start:
\begin{align*}
H_N\left(A_{R}, \pi\right) = \frac{ H\left(A_{R}, \pi\right) }{ cost\left(\pi_{A_{H}}\right) - cost\left(\pi_{A_{H} + A_{R}}^{*}\right) } . 
\end{align*}
\end{definition}

The denominator used in normalized helpfulness represents the best-case helpfulness that could be achieved for the team, given the task and domain constraints.

Another useful 
variation of helpfulness is that of \emph{relative helpfulness}, which intuitively calculates the  decrease in cost for the team to perform the task relative to the cost of the human performing the task alone~\cite{RikerLevinePhdThesis}.

\begin{definition} \label{Def:RelativeHelpfulness}
The \emph{relative helpfulness} of an agent $A_{R}$ in a plan $\pi$ is the ratio comparing the helpfulness to the best-case of the human $A_H$ performing the task by themselves:
\begin{align*}
H_R\left(A_R, \pi\right) 
&= \frac{cost(\pi_{A_H}^{*}) - cost(\pi_{A_H + A_H})}{cost(\pi_{A_H}^{*}) }
\end{align*}
\end{definition}

\section{The Cost in Helpfulness} \label{Sec:Cost}

The previous definitions of helpfulness made extensive use of the $cost(\cdot)$ function. In this section, we describe some example costs in greater detail and discuss some ramifications of different cost functions.

In the blueberry pie example, we discussed the case where $cost(\cdot)$ is the number of steps to reach the goal.  Counting the number of steps is the same as assuming that all actions have uniform cost.  Traditionally, a plan's cost is the sum of the costs of all the actions, no matter which agent performs them.  This is because \emph{automated planning is primarily concerned with problem solving more than agent contribution}, which means all optimal plans are equally ideal even if each agent's efforts are disproportional.

The notion of helpfulness is applicable to many different extensions of automated planning and execution. In temporal planning and execution problems, the \emph{makespan} (i.e., time in seconds to completion) is a valuable cost metric~\cite{DBLP:conf/aips/BentonCC12,wang2015tburton}.
Other domains may measure the cost as the amount of ``effort'' that the agents are required to exert. In multi-agent package delivery problems, there may be a human driver picking up and delivering packages as well as a semi-autonomous vehicle capable of most driving tasks. In such a case, the cost of a plan can be measured as the amount of mileage that the human drives. A helpful semi-autonomous vehicle would therefore optimize the pickup and delivery routes to reduce the human's contributions to driving. 

While it is possible to exclusively consider the costs of $A_{H}$'s actions when computing $cost(\cdot)$ for plans, 
such definitions introduce certain issues.  If $A_{H}$ and $A_{R}$ are both capable of solving the problem on their own, then normalized helpfulness is only maximized when $A_{R}$ does \emph{all the work}.  If $A_{R}$ is ever assigned a \emph{utility function to maximize their helpfulness}, then this effectively encourages the robot to act like a lackey and also encourages the human to be as lazy as possible and take advantage of their robot partner(s).  One possible solution is to ensure that doing nothing, such as performing a no-op action, has a non-$0$ cost.  Then $A_{H}$'s portion of the plan has some cost even if they do nothing, and $A_{R}$ is most helpful when all agents split the workload in tandem.  However, it is reasonable to question whether no-op should have a cost for some domain and $cost(\cdot)$ function---an agent is \emph{unlikely to expend resources such as energy} when doing nothing, but doing nothing for the duration that the other agent acts is \emph{spending resources such as time}.

It is not necessarily the case that resources alone define the cost function.  Humans have a variety of biases, emotions, preferences, and other features that will skew their perspective of performing various actions.  If a robot can reasonably model its human partner(s) to account for these aspects, then the robot can also use the information to consider more human-centric costs for being helpful.  
For example, if the human driver delivering packages prefers to drive on the freeway and dislikes driving through residential areas, then the robot can scale the action costs to enforce $cost(\text{freeway}) < cost(\text{no-op}) < cost(\text{residential})$.  It might be more helpful if the robot did not perform some tasks that would require the human to increase their cognitive load managing the robot; likewise, it would be more helpful if a robot performed tasks in which the human desires help.  Regardless of the potential loss in resource savings, \emph{acting in reasonable ways that satisfy the human can be perceived as more helpful}.

\section{Helpfulness and Uncertainty} \label{Sec:ExecutionRisk}

Thus far, we have described helpfulness in a deterministic setting. In this section, we generalize helpfulness to stochastic settings and probabilistic planning.

In many settings where a robot interacts with a human, the robot has uncertainty regarding the human's goals or plan to achieve those goals. For example: given a robot's observation that the human has obtained blueberries from the refrigerator, will the human be making a blueberry pie, a smoothie with other fruits, 
or something else? This uncertainty makes it difficult for a robot to estimate its helpfulness in real-time during interaction because the precise joint plan $\pi_{A_R + A_H}$ is uncertain. To cope with this, a robot might maintain a probability distribution over the human's goals and/or actions \cite{DBLP:conf/aaai/RamirezG10}, and thereby maintain a probability distribution over possible team plans for the task.

Given the uncertainty over team plans, helpfulness becomes a random variable. We may hence compute certain statistics, such as its expected value.

\begin{definition}
\label{Def:ExpectedHelpfulness}
Given the set of all possible human-robot team plans $\Pi$, the probability of each $\pi \in \Pi$ (denoted $\Pr(\pi)$), and the robot's set of observations so far $\mathcal{O}$, the \emph{expected helpfulness} is the expected value of relative helpfulness over the plans given those observations:
\begin{align*}
\mathbb{E}[H_R \;|\; \mathcal{O}] &= \sum_{\pi \in \Pi}{H_R(A_R, \pi) \cdot \Pr(\pi \;|\; \mathcal{O})}
\end{align*}
\end{definition}

Note that other variations of helpfulness can be used in the above (such as, for example, normalized helpfulness).

Expected helpfulness is useful for a robot to estimate how helpful it will be in the midst of executing a plan, given its uncertainty about the world. Another valuable metric is the standard deviation of helpfulness (which can be defined similarly to Definition~\ref{Def:ExpectedHelpfulness} above), a measure of the range of different helpfulness values that could result at the end of execution. As execution proceeds and the human-robot interaction unfolds, the responsive robot observes more of the human's actions $\mathcal{O}^{\prime}$, thereby updating the a posteriori probability distribution $\Pr(\pi \;|\; \mathcal{O} \cup \mathcal{O}^{\prime})$.

One risk-aware executive for controlling a robot is \riker~\cite{RikerLevinePhdThesis}. \riker explicitly maintains a probability distribution over the human's choices (such as actions or goals) in contingent, temporally-flexible team plans. \riker chooses the robot's actions in a risk-bounded manner: it aims to ensure that the probability that the team will fail their task is at most $\Delta$, where $0 \leq \Delta \leq 1$. By tuning $\Delta$, qualitatively different behaviors result as the robot becomes more or less risk-averse. Low values of $\Delta$ make the robot more risk-averse (less likely to take risk) whereas higher values cause the robot to boldly take greater risks.

As the experimental results in both Section~\ref{results.exp} and the spelling Blocksworld domain in \citet{RikerLevinePhdThesis} illustrate, there is a fundamental trade-off between risk and expected helpfulness. A risk-averse robot (such as when $\Delta$ is low) will be less willing to take any actions that may jeopardize execution. As such, a risk-averse robot's safest action is often to do nothing unless it is positive it can help. This results in team costs being higher, sometimes even approaching the cost of the human acting alone.  Hence helpfulness values will be lower. As the risk bound is loosened and $\Delta$ increases, the robot will tolerate more risk and act sooner, even if it is not completely certain of the human's intentions. In this case, helpfulness values tend to be greater for cases when the team succeeds, but lower in the cases where the robot incorrectly assumes the goal and selects incorrect actions that often require backtracking.

\section{Helpfulness During Planning\label{planningEx}}

Rather than passively measure helpfulness as a post-hoc analysis, there are also potential uses to measuring it \emph{actively during interaction}.  Specifically, when $A_{R}$ intends to independently define their own goals with respect to $A_{H}$'s, increasing helpfulness is ideal for assistive interaction, and decreasing helpfulness is ideal for adversarial interaction.  In the case of collaborative interaction where $A_{R}$ and $A_{H}$ share a common goal throughout the experience, it is ideal to increase helpfulness as an assistive behavior for teammates' subtasks and maintain non-decreasing helpfulness when working on independent subtasks.  These motivate the use of helpfulness to \emph{guide $A_{R}$'s decision making process} throughout their interaction with $A_{H}$.

Typically, the type of interaction exclusively plays a role in $A_{R}$'s goal generation \cite{DBLP:conf/aaai/FreedmanZ17}, and the planner then finds any solution to this goal from the initial state.  Although this sequence of actions satisfies the goal upon completion, there is no guarantee that \emph{the interaction type is acknowledged throughout the responsive plan's execution}.  Research in legible planning \cite{Dragan:2014:IHO:2678082.2678168_disambiguousReaching,DBLP:conf/aaai/KulkarniSK19,mz_pair2020} shows that optimal-cost plans to the goal are often less interpretable for communicating one's intents, and additional criteria need to influence the planner to account for a higher-cost, but more legible plan.  In most cases, these criteria adjust the heuristic or constraints to reshape the search progression because the \emph{path to the goal matters as much as the goal itself}.

In place of legibility, it is ideal for the robot's plan $\pi_{A_{R} \rightarrow A_{H}}$ to be assitive, independent, or adversarial throughout.  A planner's search state should not be considered `near a goal state' simply because its cost is lower, but also because it is a state that exhibits some degree of helpfulness with respect to the interaction type.  In assistive interactions, helpfulness can serve as a tie-breaking strategy between states on the frontier with the same lowest expected cost to the goal---the 
state whose current plan has greatest helpfulness should be considered first.  For independent interactions, helpfulness might not matter as much as long as (1) $A_{R}$ does not interfere with $A_{H}$'s (possibly inferred) goals and (2) $A_{R}$ does not go too far out of the way from achieving their own goals.  To ensure that $A_{R}$'s plan is minimally invasive, helpfulness can serve as a tie-breaking strategy by selecting the state whose current plan's helpfulness has the least absolute value.  As the final extreme, adversarial interactions should prioritize tie-breaking towards states whose current plan has the least helpfulness.  To ensure that helpfulness is not ignored when tie-breaking is unnecessary during search, we can adjust the heuristic to account for both expected cost and helpfulness as either a linear combination \cite{Hansen07} or a lexicographic preference for multiple objectives \cite{wz_ijcai15,yfw_aaai2015}.

One caveat to computing helpfulness during planning, as opposed to afterwards, is the need to \emph{approximate helpfulness} given the plan up to a search state. 
As a post-interaction computation, helpfulness is easy to compute because the (possibly inferred) goal is known when searching for $A_{H}$'s single-agent plan \emph{and} we know what actions $A_{H}$ took to accomplish their goal.  In a centralized multi-agent setting, the planner should know the shared goal $G$ and at least have access to a heuristic that involves all agents present $h_{joint}$ in addition to individual heuristics for each agent $h_{A_{\cdot}}$.  Such planners can estimate helpfulness from any search state $s$:
\begin{align}\label{eq:helpheur_joint}
    h_{help}^{+} = h_{A_{H}}\left(s, G\right) - h_{joint}\left(s, G\right) .
\end{align}
If the planner uses a heuristic search technique such as A$^{*}$ and prefers to only sort candidate states by potential helpfulness, then we can extend this to compute a priority value for the frontier using initial state $I$ and path-so-far cost functions $g_{joint}$ and $g_{A_{\cdot}}$ 
(which can be as expensive to compute as a new search problem for $A_{H}$ finding a path from $I$ to $s$):
\begin{align*}
    f_{help}^{+} = & g_{help} + h^{+}_{help} \\
    g_{help} = & g_{A_{H}}\left(I, s\right) - g_{joint}\left(I, s\right)
\end{align*}
On the other hand, a decentralized agent $A_{R}$ is not guaranteed to know $A_{H}$'s true goal or future actions; the planner must approximate both of these in the search.  The computational efforts to predict $A_{H}$'s remaining actions could be high unless we make trivializing assumptions such as $A_{R}$ no longer acting, which would reduce predicting the cost of $A_{H}$'s remaining actions to the traditional single-agent cost heuristic $h_{A_{H}}$ from $s$ and effectively set helpfulness to $0$.  As this is not ideal and counteracts the entire purpose of heuristic search, we instead replace $G$ in Equation~\ref{eq:helpheur_joint} with some inferred goal conditions $\widehat{G}$ and apply scalar $\alpha \in \left[0,1\right]$ for the proportion of the joint plan that we expect $A_{H}$ to perform: 
\begin{align}
h_{help}^{\rightarrow} = h_{A_{H}}\left(s, \widehat{G}\right) - \alpha \cdot h_{joint}\left(s, \widehat{G}\right) .
\end{align}
If the planner uses heuristic search and prefers to only prioritize states by their helpfulness, then the above equations almost identically apply.  $g_{help}$ does not change because the path-so-far from $I$ to $s$ is already complete.  $f_{help}^{\rightarrow}$ simply replaces $h_{help}^{+}$ from $f_{help}^{+}$ with $h_{help}^{\rightarrow}$.

\section{Preliminary Results\label{results}}
Due to the current safety concerns of performing human subjects experiments during a global pandemic, we investigate the use of helpfulness and its proposed variations with simulated AI-controlled agents.  The human agent $A_{H}$ uses simple heuristic search algorithms and the robot agent $A_{R}$ employs either the \riker \cite{RikerLevinePhdThesis} executive or a responsive-planning implementation of the \pretcil framework \cite{freedmanthesis}.

\subsection{Domain: Foodworld\label{results.domain}}
To follow our running example of preparing a blueberry pie, we use a food-assembly-themed domain that is effectively the traditional Blocksworld domain. 
Rather than stacking blocks with inscribed letters to spell words (when read from top-to-bottom), the \emph{Foodworld} domain stacks food items and utensils to create dishes.  For example, making the blueberry pie requires the following stack from bottom-to-top: pan, butter, dough, blueberries, and sugar. 

Each agent may hold up to one item at a time via pick-and-place actions, and they can only pick up items on the top of each stack---likewise, placing an item puts it on top of the chosen stack.  The action set for the human alone $A_{H}$ is picking up an item, placing an item, or doing nothing; all actions have uniform cost.  The action set for the centralized team $A_{R}+A_{H}$ extends this for parallel execution where both agents may act simultaneously; each human-robot pair of actions has uniform cost.  By this design, our cost function is defined relative to step time.

For the experiments that take place in the Foodworld domain, we had twelve food items and utensils and up to six possible dishes.  Dishes required as few as three and as many as five of the twelve items available.  To illustrate the impact of the initial state on solving a problem, we created two different initial states: the organized kitchen has almost all items alone in a single stack (easy access to most items), and the cluttered kitchen condenses all the items into six vertical stacks containing up to three items (some items require more effort to access).

\subsection{Helpfulness Computations \label{results.exp}}
Table~\ref{tab:results.exp.helpvals} lists the costs of optimal plans (we simply ran A$^{*}$ Search) for single- and joint-agent scenarios in our Foodworld problem instances (two initial states $\times$ six goals).  In general, the cost to prepare a dish increases as more items are required, and the cost is also greater when preparing the dish in the cluttered kitchen.

The helpfulness is positive, which tells us that the robot always reduced the time taken to prepare a dish.  For comparison, a vacuum robot without any limbs cannot perform these pick-and-place actions so that $cost\left(\pi_{A_{H}}^{*}\right) = cost\left(\pi_{A_{R}+A_{H}}^{*}\right)$ and $H = 0$.  However, these integer measurements lack standardization as the optimal plan costs vary greatly between problem instances.  The normalized helpfulness provides context to how these helpfulness values compare between problem instances---because all these plans are optimal, we know that the robot has reduced the plans' costs \emph{as much as possible} for a consistent value of $1$.  On the other hand, the relative helpfulness portrays the portion of the human's overall effort that was relieved.  Unsurprisingly, the robot's relative helpfulness was greater in the cluttered kitchen per dish; the lack of organization provides more opportunities to unstack items so that the human can access them more easily without rummaging through the stacks.

\begin{figure}
    \centering
    \includegraphics[scale=0.6]{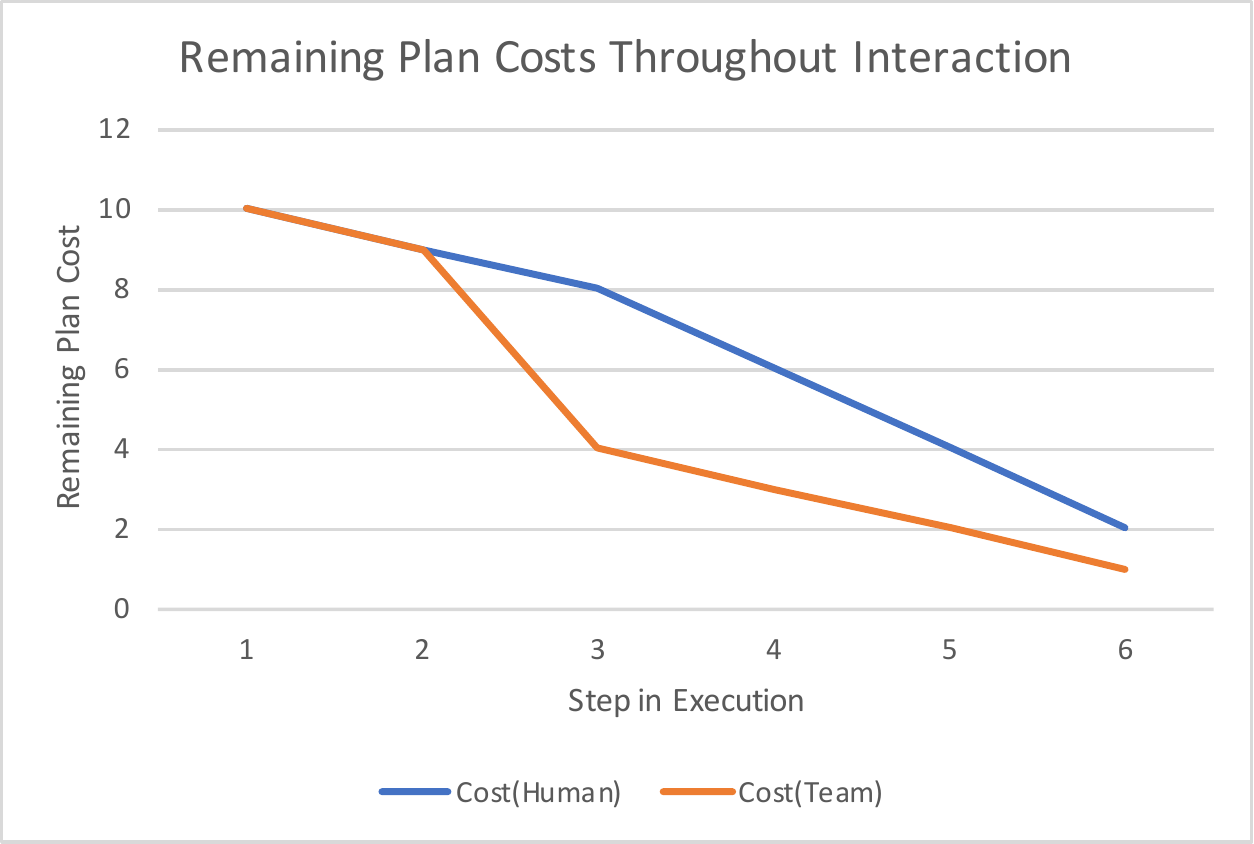}
    \caption{Plan costs to make a blueberry pie, starting from each state throughout the interaction between $A_{H}$ and $A_{R}$.}
    \label{fig:results.exp.remainplancosts}
\end{figure}

These interpretations empirically hold for decentralized agents as well.  Figure~\ref{fig:results.exp.remainplancosts} plots the remaining plan costs as two agents interact in the organized kitchen to make a blueberry pie.  The human agent simply runs A$^{*}$ search in the current state, but the robot employs an implementation of the \pretcil framework that integrates probabilistic recognition as planning \cite{DBLP:conf/aaai/RamirezG10} with heuristic search.  The \pretcil framework is an architecture for closed-loop interaction that observes other agents, infers those agents' goals, generates its own intermediate goals based on these inferences, plans for the intermediate goals, and continues to monitor the interaction for more observations while executing the plan.  In this case, the robot observes without acting for the first two time steps so that the human does all the work.  After two time steps, the robot identifies that the human is making a blueberry pie and assists with the task.  Despite the lost time, both agents can finish the blueberry pie together in four time steps so that it still takes six steps overall; this was the optimal cost for the joint plan.  Therefore, the helpfulness changes from zero to four after two time steps, and the normalized and relative helpfulness instantly jump from $0$ to $1$ and from $0$ to $0.4$ respectively.

\begin{table*}
\centering
\caption{Helpfulness Evaluations in Foodworld (Joint Plan)\label{tab:results.exp.helpvals}}
\begin{tabular}{|l|c|c||c|c|c||c|c||c|c|c|}
\hline
     & \multicolumn{5}{|c|}{Organized Kitchen} & \multicolumn{5}{|c|}{Cluttered Kitchen} \\
\hline
\hline
Dish (\# Items in Stack) & Cost   & Cost  & $H$ & $H_{N}$ & $H_{R}$ & Cost   & Cost  & $H$ & $H_{N}$ & $H_{R}$ \\
     & ($\pi_{A_{H}}^{*}$) & ($\pi_{A_{R}+A_{H}}^{*}$) &     &         &         & ($\pi_{A_{H}}^{*}$)\vspace{3pt} & ($\pi_{A_{R}+A_{H}}^{*}$) &     &         &          \\
\hline
Sugar Cookie (3) & 6 & 4 & 2 & 1 & $0.\overline{3}$ & 10 & 6 & 4 & 1 & 0.4 \\
Blueberry Pie (5) & 10 & 6 & 4 & 1 & 0.4 & 16 & 8 & 8 & 1 & 0.5 \\
Fudge (4) & 8 & 5 & 3 & 1 & 0.375 & 12 & 6 & 6 & 1 & 0.5 \\
Jelly Donut (3) & 6 & 4 & 2 & 1 & $0.\overline{3}$ & 14 & 8 & 6 & 1 & 0.429 \\
Choco-Chip Cookie (3) & 6 & 4 & 2 & 1 & $0.\overline{3}$ & 12 & 6 & 6 & 1 & 0.5 \\
Cake (5) & 12 & 6 & 6 & 1 & 0.5 & 12 & 6 & 6 & 1 & 0.5 \\
\hline
\end{tabular}
\end{table*}

\subsection{Expected Helpfulness Results}

\begin{figure}
    \centering
    \includegraphics[width=\linewidth]{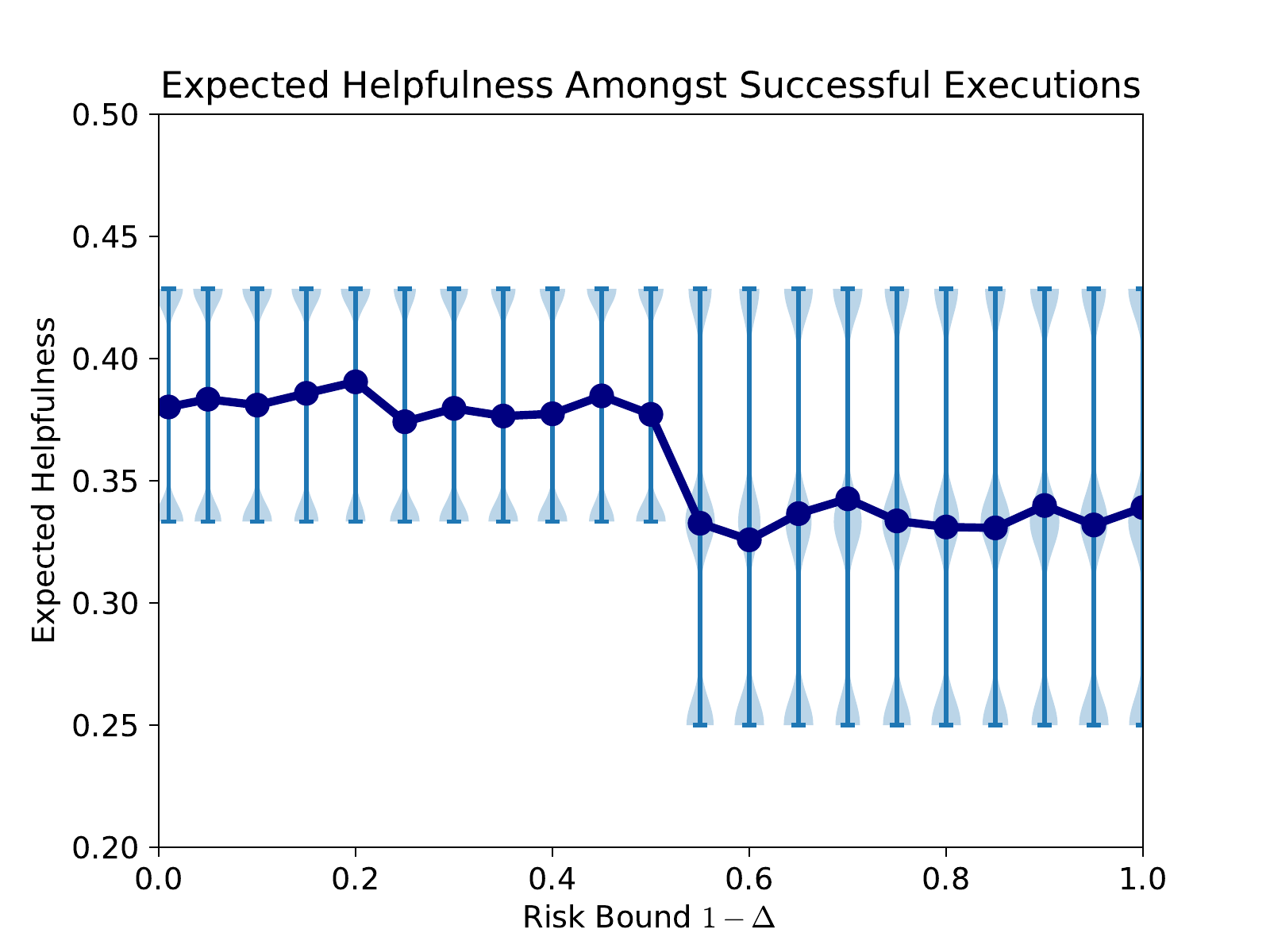}
    \caption{Expected helpfulness $\mathbb{E}[H_R \;|\; \mathcal{O}]$ as a function of \riker's risk bound $\Delta$. As \riker becomes more conservative (towards right), its expected helpfulness decreases.}
    \label{fig:ExpectedHelpfulness}
\end{figure}

In this section, we illustrate the trade-off between expected helpfulness and risk using \riker. We encoded a breakfast-making scenario between a human and a robot as a contingent, temporally-flexible plan (i.e., as a probabilistic temporal plan network~\cite{DBLP:conf/aips/SantanaW14}). In this plan, the team prepares either oatmeal, cold cereal with milk, or waffles. For each breakfast, the contingent plan encodes two methods of achieving the goals: one in which the robot acts immediately, and the other in which the robot begins acting after a short delay (and is able to observe some of the human's actions first and hence infer intent). We also encoded a PDDL model~\cite{PDDL} so that \riker could derive causal links in order to find relations between choices in the contingent plan. We lastly encoded a Bayesian network to describe the probability distribution over the human's intentions in this plan, with all breakfasts being equally likely. This model is analogous to the collaborative spelling blocksworld model that \citet{RikerLevinePhdThesis} describes.

Execution of this breakfast-making scenario was simulated thousands of times with \riker to show the relationship between expected helpfulness and aversion to risk. During these simulations, \riker's risk tolerance $\Delta$ was varied. For each value of $\Delta$, execution was simulated 50 times and the corresponding helpfulness of the resulting execution was measured. The expected helpfulness $\mathbb{E}[H_R \;|\; \mathcal{O}]$ was then computed by taking the mean helpfulness (hence, this was a Monte Carlo simulation). In total, 2100 simulations were performed. 

The results are visualized in Figure~\ref{fig:ExpectedHelpfulness}. \riker's risk bound $\Delta$ is plotted along the $x$-axis, and the expected helpfulness $\mathbb{E}[H_R \;|\; \mathcal{O}]$ is shown on the $y$-axis. Light blue bands illustrate the relative distribution of discrete helpfulness values, and the dark blue curve illustrates $\mathbb{E}[H_R \;|\; \mathcal{O}]$.  Our key observation is that, as \riker becomes more cautious and more risk-averse (towards the right), we see that the expected helpfulness decreases (this is the drop that occurs near $\Delta = 0.5$). While the expected helpfulness was greater for cases where \riker took more risk, execution was also more likely to fail in these cases (for instance, by the robot guessing the wrong breakfast and taking action prematurely). The simulations described here assume that execution ends when such failures occur, and calculate $\mathbb{E}[H_R \;|\; \mathcal{O}]$ up to that point.
Alternatively, it is possible to assume that execution continues after a failure -- which could possibly result in a lower helpfulness (but no failure) for those runs.

\section{Conclusions and Future Work\label{conclusions}}

The abstract nature of \emph{helping} is generally evaluated more holistically compared to the corresponding tasks that have concrete goal conditions.  It is often easy to be successful with respect to the interaction's description, but the human could deem the robot's \emph{level of success} lackluster.  Poor evaluations could impact the human's trust.  For example:
\begin{itemize}
\item A robot carries \emph{only} a single bag of pretzels to the kitchen to help, but the human carries the remaining groceries including heavy milk cartons and canned food.
\item A robot treads around the carpet's perimeter to avoid getting in the way of a human vacuuming the carpet, but trips over the cord and disconnects the vacuum from the wall outlet.  So the human has to stop and plug it back in. 
\item A robot hides all the musical instruments and the television remote in another room away from noisy children, which only delays the children from creating noise until they find the hiding spot.
\end{itemize}
We introduced helpfulness and some variations as a means of measuring the degree to which a robot 
reduced a human's 
efforts, and ideal values for this measurement depend on the type of interaction.  This creates a consistent, quantified evaluation of a robot's involvement in interactions with people.

It is likely that helpfulness can play a deeper role in the decision-making process so that responsive plans adhere to helping for the entirety of the interaction, rather than be a byproduct of achieving each intermediate concrete task.  We explored this potential through both expected helpfulness and a helpfulness heuristic.

Future work includes extending the efforts above to consider additional variations and interpretations of helpfulness, impacts and properties of various cost function definitions, and applications of helpfulness throughout the interactive experience.  In addition to the theoretical perspective, we intend to explore the practical side of helpfulness to evaluate decision making in human-robot interaction.  
We would like to perform similar experiments to those in this \doctype with actual people rather than simulated 
agents.  Furthermore, as people do not always perceive things computationally, it is important to compare how people evaluate a robot's helpfulness (via Likert scales and other metrics) to our proposed quantified metric.  If we expect humans to use various forms of helpfulness to assess their robot partners, then they need to match humans' expectations in order to be informative \emph{and helpful}.

\section*{Acknowledgments}
The authors would like to thank the anonymous reviewers for their useful feedback and insightful perspectives.

\begin{small}
\bibliographystyle{aaai} \bibliography{biblio}
\end{small}

\end{document}